\documentclass[letterpaper]{article} %
\usepackage{aaai24}  %
\usepackage{times}  %
\usepackage{helvet}  %
\usepackage{courier}  %
\usepackage[hyphens]{url}  %
\usepackage{graphicx} %
\usepackage{natbib}  %
\usepackage{caption} %
\usepackage{algorithm}
\usepackage{algorithmic}

\usepackage{amsmath} 
\usepackage{booktabs}
\usepackage{makecell}
\usepackage{bm}
\usepackage{amsfonts}

\usepackage{newfloat}
\usepackage{listings}
\DeclareCaptionStyle{ruled}{labelfont=normalfont,labelsep=colon,strut=off} %
\floatstyle{ruled}
\newfloat{listing}{tb}{lst}{}
\floatname{listing}{Listing}
\nocopyright %

\title{Geometry-aware Line Graph Transformer Pre-training for \\
Molecular Property Prediction}
\author{
    Peizhen Bai,\thanks{This is a preprint.}
    Xianyuan Liu,
    Haiping Lu
}
\affiliations{
    Department of Computer Science, University of Sheffield

    pbai2@sheffield.ac.uk, 
    xianyuan.liu@outlook.com,
    h.lu@sheffield.ac.uk
}

\usepackage{bibentry}

\begin{document}

\maketitle

\begin{abstract}
Molecular property prediction with deep learning has gained much attention over the past years. Owing to the scarcity of labeled molecules, there has been growing interest in self-supervised learning methods that learn generalizable molecular representations from unlabeled data. Molecules are typically treated as 2D topological graphs in modeling, but it has been discovered that their 3D geometry is of great importance in determining molecular functionalities. In this paper, we propose the Geometry-aware line graph transformer (Galformer) pre-training, a novel self-supervised learning framework that aims to enhance molecular representation learning with 2D and 3D modalities. Specifically, we first design a dual-modality line graph transformer backbone to encode the topological and geometric information of a molecule. The designed backbone incorporates effective structural encodings to capture graph structures from both modalities. Then we devise two complementary pre-training tasks at the inter and intra-modality levels. These tasks provide properly supervised information and extract discriminative 2D and 3D knowledge from unlabeled molecules. Finally, we evaluate Galformer against six state-of-the-art baselines on twelve property prediction benchmarks via downstream fine-tuning. Experimental results show that Galformer consistently outperforms all baselines on both classification and regression tasks, demonstrating its effectiveness.
\end{abstract}

\section{Introduction}

Predicting molecular properties plays a crucial role in drug discovery and computational chemistry. With the advancement of deep learning, many recent works exploit learning generalizable molecular representations by the self-supervised learning (SSL) paradigm, where a model is pre-trained with large-scale unlabeled molecules, and then fine-tuned to downstream property predictions with limited labeled data. SSL has shown its efficacy in capturing molecular information and improving predictive performance \cite{hu2020pretraining, zhang2021motif, xu2021self}.

In recent years, most molecular self-supervised learning methods have utilized graph neural networks (GNNs) for encoding 2D molecular graphs, where atoms are represented as nodes and chemical bonds as edges. Depending on the message-passing mechanism, GNNs can effectively preserve topological information in molecules. One line of these works develops graph contrastive learning-based methods, which adopt data augmentation strategies to generate correlated graph pairs from the same molecule \cite{You2020GraphCL, suresh2021adversarial}. However, these works have only considered the 2D modality of molecular data, despite its multimodal nature. More recently, there has been a growing trend towards integrating 3D geometric information into 2D molecular representation learning, and designing new contrastive methods across modalities \cite{Strk20213DII, zhu2022unified, liu2022pretraining}. The incorporation of 3D geometry can provide rich energy knowledge that is vital in determining molecular functionalities.

\begin{figure*}[htbp]
    \centering
    \includegraphics[width=0.8\textwidth]
    {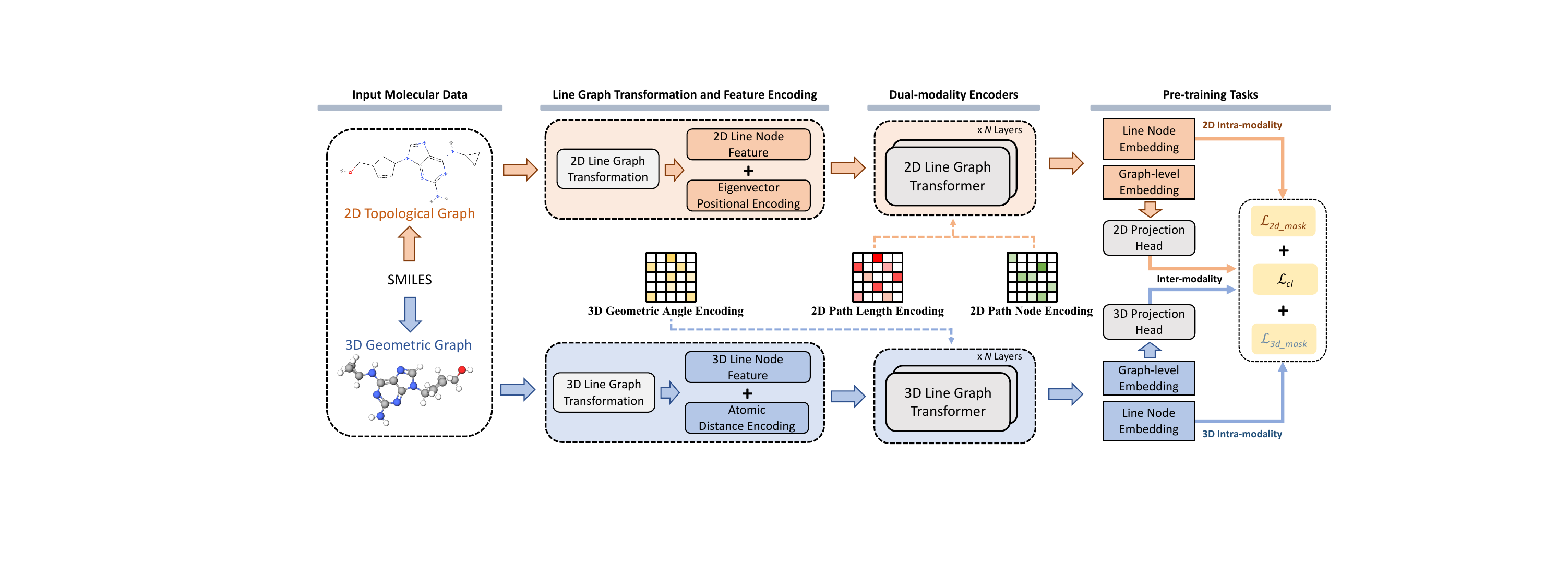}
    \caption{Overview of the Galformer pre-training framework on 2D and 3D molecular graphs.}
    \label{main_flow}
    \vspace{-1.5em}
\end{figure*}

Despite the fruitful developments, two challenges remain in 2D-3D molecular self-supervised learning: limited model capacity and incomplete pre-training tasks. Firstly, previous studies primarily employ existing 2D and 3D GNNs as backbone networks for pre-training, which have consistently suffered from the over-smoothing problem and expressive power limitation \cite{Xu2018HowPA}. These issues limit the model capacity to capture rich multimodal information and learn expressive representations from molecular data. Then most dual-modality molecular pre-training methods only focus on learning inter-modality relation between 2D and 3D molecular structures, without simultaneous consideration of intra-modality relation. The intra-modality relation within each modality is complementary to the inter-modality relation. For instance, each 2D molecular representation should not only obtain geometric knowledge from its associated 3D modality, but also preserve the topological information within the 2D structure. By considering both inter- and intra-modality knowledge extraction in a unified molecular SSL framework, the pre-trained model is more robust and generalizable to various downstream tasks.

To tackle the two challenges, we propose the \textbf{G}eometry-\textbf{a}ware \textbf{l}ine graph trans\textbf{former} (Galformer). It is a unified 2D-3D pre-training framework to learn generalizable molecular representations. Figure \ref{main_flow} presents the overview of Galformer. We first transform the input 2D and 3D molecular graphs into their line graphs (i.e. edge-adjacency graphs). This transformation is to preserve the inherent adjacency information in molecules while emphasizing structural information at finer granularity levels, such as node-pair distances and edge-pair angles. Subsequently, we design a dual-modality line graph transformer architecture to model both 2D and 3D molecular information flow simultaneously. To effectively consider the inter- and intra-modality relations in SSL, we design two complementary pre-training tasks: masked line node prediction and dual-view contrastive learning. At the intra-modality level, we randomly mask a proportion of 2D and 3D line nodes, and then predict their types from the context embedding within the individual modality. Owing to the property of the line graph, this strategy considers local atom-pair, bond, and angle information being masked, thus capturing richer structural patterns in molecules. At the inter-modality level, we apply contrastive learning to maximize the mutual information between the graph-level 2D and 3D molecular representations. The contrastive pre-training task bridges the modality gap and is complementary to inter-modality mask learning. The pre-trained 2D encoder incorporates rich 3D geometric knowledge to improve downstream property predictions. Our contributions are three-fold:

\begin{itemize}
    \item We propose a dual-modality line graph transformer architecture to encode both 2D topological and 3D geometric information of molecules. By considering molecular structures as graph inductive bias into the transformer, the designed backbone can extract discriminative knowledge across modalities.
    \item We design two complementary pre-training tasks: masked line node prediction at the inter-modality level and dual-view contrastive learning at the intra-modality level. Both contribute to generating more robust and generalizable molecular representations.
    \item We evaluate Galformer performance on twelve downstream molecular property datasets, including both classification and regression tasks. The experimental results show its superiority over six state-of-the-art baselines.
\end{itemize}

\section{Related Work}
\textbf{Molecular Representation Learning.} As the critical foundation of property prediction, many studies have been devoted to improving molecular representation learning. In recent years, the emergence of graph neural networks has allowed for more intuitive and efficient learning of molecular representations. \citet{gilmer2017neural} first proposes to use a message-passing framework to capture atomic interactions in molecular graphs, and then \citet{yang2019analyzing} extends the framework by considering the directional bond interactions. Nevertheless, these methods still face challenges in handling the geometry of molecules and capturing long-range dependencies, which has led to the exploration of transformer-based architectures to preserve graph structures and improve the expressiveness of learned molecular representations \cite{ying2021transformers, pyzer2022accelerating, luo2022clear}.

\noindent\textbf{Molecular Self-supervised Learning.} Self-supervised learning has achieved fruitful progress in the area of molecular representation learning, leveraging the potential of unlabeled molecular data. Many pre-training tasks have been developed and can be broadly classified into contrastive methods and generative methods \cite{liu2022graph, jiao20223d, feng2022mgmae}. The principal objective of contrastive methods is to maximize the mutual information between the augmented perspectives from the same molecular graph. These methods suggest different augmentation strategies to create semantically similar perspectives \cite{You2020GraphCL, Wang2021MolecularCL, Sun2021MoCLDM}. In contrast, generative methods consider reconstructing the information of a single molecule at different levels, or creating certain pseudo labels for pre-training \cite{rong2020self, Li2022KPGTKP}. \citet{zhang2021motif} proposes to learn molecular representations from both node-level and graph-level fragmentations, which predicts masked atom/bond-based attributes and reconstructs motif-based graph trees.

\section{Preliminaries}

\textbf{Definition 1 (2D Topological graph).} Given a 2D molecule, its topological graph can be defined as $\mathcal{G}^{2d} = (\mathcal{V}^{2d}, \mathcal{E}^{2d})$, where each node $u \in \mathcal{V}^{2d}$ denotes an atom and each edge $(u, v) \in \mathcal{E}$ represents a chemical bond between node (atom) $u$ and $v$. Meanwhile, we initialize the node feature with atom attributes as $\mathbf{h}^{2d}_u \in \mathbb{R}^{\Theta_{v}}$ and edge feature with chemical bond attributes as $\mathbf{e}^{2d}_{uv} \in \mathbb{R}^{\Theta_{e}}$, where $\Theta_{v}$ and $\Theta_{e}$ denotes the dimensions of initial node and edge features, respectively.

\noindent\textbf{Definition 2 (3D Geometric Graph).} A geometric graph of 3D molecule can be represented as $\mathcal{G}^{3d} = (\mathcal{V}^{3d}, \mathcal{E}^{3d}, \mathcal{A}^{3d})$, where $(u, v, w) \in \mathcal{A}$ denotes the geometric angle between edge $(u, v)$ and $(v, w)$. Similar to previous studies \cite{li2022geomgcl, fang2022geometry, feng2022mgmae}, we consider the atomic distance $d_{uv} > 0$ and angle $\theta_{uvw} \in [0, \pi]$ as invariant spatial features in $\mathcal{G}^{3d}$, regardless of how the same molecular conformation rotates or translates in 3D space. To emphasize the geometric information, each 3D node feature is denoted as a one-hot vector $\mathbf{h}^{3d}_u \in \mathbb{R}^{\Theta_{t}}$, where ${\Theta_{t}}$ is the number of atomic types without additional topological attributes.

\noindent\textbf{Problem Formulation.} Given a set of unlabeled molecules $\mathcal{M} = \{\mathcal{G}^{2d}_{i}, \mathcal{G}^{3d}_{i}\}_{i=1}^{|\mathcal{M}|}$, where each molecule $M_i \in \mathcal{M}$ has its 2D topological graph $\mathcal{G}^{2d}_i$ and 3D geometric graph $\mathcal{G}^{3d}_i$. The study aims to pre-train a dual-modality SSL model that generates discriminative molecular representations and then adapts to molecular property prediction by fine-tuning. Due to the scarcity of the 3D conformers in downstream tasks, following the paradigm of previous works \cite{liu2022pretraining, Strk20213DII}, we regard the knowledge of 3D geometry as privileged information only used during pre-training. Then the pre-trained 2D encoder is subsequently fine-tuned for downstream property prediction tasks, with only 2D topological graphs available.

\vspace{-0.5em}
\section{Galformer: The Proposed Framework}
In this section, we present the details of our proposed pre-training framework, termed Galformer. Firstly, we describe the procedure for transforming 2D and 3D molecular graphs into their corresponding line graphs. Then we elaborate on the design of dual-modality line graph transformers that can effectively encode both topological and geometric information. Finally, we introduce the pre-training strategy that enables learning of inter-modality and intra-modality relations.

\subsection{Molecular Line Graphs}
\begin{figure}[t]
    \centering
    \includegraphics[width=0.42\textwidth]{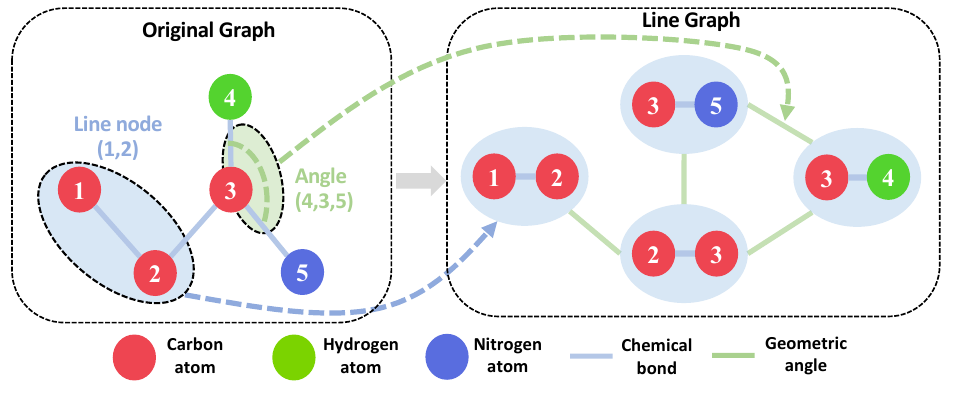}
    \caption{An illustration of transforming a molecular graph to its line graph.}
    \label{line_graph}
    \vspace{-1.5em}
\end{figure}

\subsubsection{Topological Line Graph}
Given a 2D topological graph $\mathcal{G}^{2d} = (\mathcal{V}^{2d}, \mathcal{E}^{2d})$, we can transform it into a topological line graph $\mathcal{\hat{G}}^{2d} = (\mathcal{\hat{V}}^{2d}, \mathcal{\hat{E}}^{2d})$ following the two steps: (i) For each edge in $\mathcal{G}^{2d}$, create a node in $\mathcal{\hat{G}}^{2d}$. This step involves mapping the edges in the original graph to the nodes in the line graph. (ii) For every pair of edges in $\mathcal{G}^{2d}$ that share a common node, create an edge between their corresponding nodes in $\mathcal{\hat{G}}^{2d}$. This step involves making connections between the nodes in the line graph. After the transformation, we can initialize each node feature in $\mathcal{\hat{G}}^{2d}$ with the atom and bond attributes as follows:

\begin{equation}
    \mathbf{n}^{2d}_{uv}= \mathrm{Concat}(\mathbf{W}_g\mathbf{h}_u^{2d} + \mathbf{W}_g\mathbf{h}_v^{2d}, \mathbf{W}_{e}\mathbf{e}^{2d}_{uv}),
\end{equation}

\noindent where $\mathbf{W}_g \in \mathbb{R}^{\frac{\Theta_{\hat{v}}}{2} \times \Theta_{v}}$ and $\mathbf{W}_e \in \mathbb{R}^{\frac{\Theta_{\hat{v}}}{2} \times \Theta_{e}}$ are learnable projection matrices. To avoid ambiguity and simplify notations, we denote $\mathcal{\hat{V}}^{2d} = \{\hat{v}_i\}_{i=1}^{|\mathcal{\hat{V}}^{2d}|}$ as the set of nodes in $\mathcal{\hat{G}}^{2d}$, where ${|\hat{V}^{2d}}|$ is equivalent to the number of edges $|\mathcal{E}^{2d}|$ by the definition of line graph.

\subsubsection{Geometric Line Graph.} A geometric line graph $\mathcal{\hat{G}}^{3d} = (\mathcal{\hat{V}}^{3d}, \mathcal{\hat{E}}^{3d})$ can be derived from its corresponding 3D geometric graph $\mathcal{G}^{3d}$ by the similar transformation steps above. In contrast, we emphasize integrating the invariant spatial features within the line graph framework. Specifically, we use Gaussian Basis Kernel function \cite{scholkopf1997comparing} to map the scalar atomic distances and angles into high-dimensional vectors, which can be written as:

{\footnotesize
\begin{equation}
    \mathbf{d}_{uv} = \underset{m=1}{\overset{M}{\frown}}\left(-\frac{1}{\sqrt{2\pi}|\sigma^m_d|}\mathrm{exp}\left(-\frac{1}{2}\left(\frac{\alpha^{t}_{d}d_{uv} + \beta^{t}_{d} - \mu^m_d}{|\sigma^m_d|}\right)^2\right)\right),
\end{equation}
}

{\footnotesize
\begin{equation}
    \bm{\theta}_{uvw} = \underset{m=1}{\overset{M}{\frown}}\left(-\frac{1}{\sqrt{2\pi}|\sigma^m_\theta|}\mathrm{exp}\left(-\frac{1}{2}\left(\frac{\alpha^{t}_{\theta}\theta_{uvw} + \beta^{t}_{\theta} - \mu^m_\theta}{|\sigma^m_\theta|}\right)^2\right)\right),
\end{equation}
}

\noindent where ${\frown}$ denotes the concatenation operator over multiple scalars and $M$ is the number of Gaussian Basis kernels. ($\sigma^m_d$, $\mu^m_d$) and ($\sigma^m_\theta$, $\mu^m_\theta$) are the $m$-th learnable kernel center and scaling factor for the atomic distance and angle, respectively. ($\alpha^t_d$, $\beta^t_d$) and ($\alpha^t_\theta$, $\beta^t_\theta$) are learnable scalars but indexed by the type of atomic pair in $(u, v)$ and edge pair in $(u, v, w)$, respectively. The mapped pair-wise encodings can preserve invariant spatial features and adapt to further deep modeling. Next, the node and edge features in $\hat{G}^{3d}$ are defined:

\begin{equation}
    \label{atomic_dis}
    \mathbf{n}^{3d}_{uv}= \mathrm{Concat}(\mathbf{W}_l\mathbf{h}_u^{3d} + \mathbf{W}_l\mathbf{h}_v^{3d}, \mathbf{W}_{d}\mathbf{d}_{uv}),
\end{equation}

\begin{equation}
    \label{angle}
    \mathbf{e}_{uvw}^{3d} = \mathbf{W_p}\mathbf{\bm{\theta}_{uvw}} 
\end{equation}
where $\mathbf{W}_l \in \mathbb{R}^{\frac{\Theta_{\hat{v}}}{2} \times \Theta_{t}}$, $\mathbf{W}_d \in \mathbb{R}^{\frac{\Theta_{\hat{v}}}{2} \times M}$ and $\mathbf{W}_p \in \mathbb{R}^{\Theta_{\hat{e}} \times M}$ are learnable projection matrices. By definition, the edges in the line graph represent the pairs of edges in the original graph sharing a common node (i.e. adjacent edges). The relationship derives that each line edge corresponds to a unique angle in the original graph. Therefore, we can adopt the projected angle vector $\mathbf{e}^{3d}_{uvw}$ as edge feature in $\hat{G}^{3d}$. Finally, we define $\mathcal{\hat{V}}^{3d} = \{\hat{v}_i\}_{i=1}^{|\mathcal{\hat{V}}^{3d}|}$ as the set of nodes and $\mathcal{\hat{E}}^{3d} = \{\hat{e}_{ij}\}_{i,j \in [1, {|\mathcal{\hat{V}}^{3d}|]}}$ as the set of edges in $\mathcal{\hat{G}}^{3d}$. Figure \ref{line_graph} illustrates the process of line graph transformation.

\subsection{Dual-modality Line Graph Transformers}
\begin{figure}[t]
    \centering
    \includegraphics[width=0.42\textwidth]{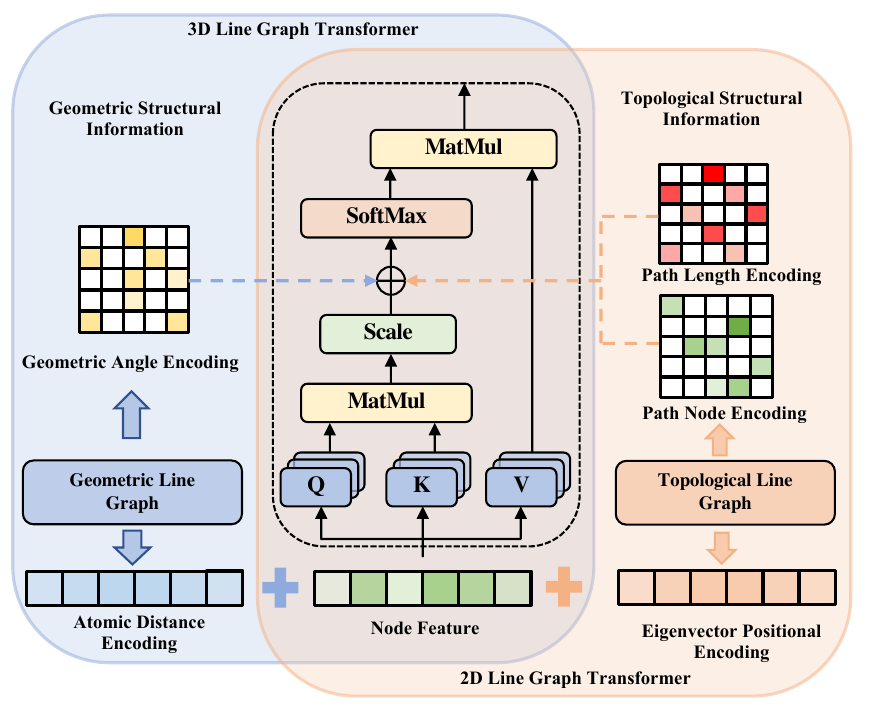}
    \caption{An illustration of the dual-modality line graph transformer architecture.} 
    \label{transformer}
    \vspace{-1.5em}
\end{figure}

Although transformer architecture has demonstrated remarkable performance in various tasks involving sequence data, its direct application to graph-structured data still remains challenging. The primary obstacle lies in the fact that the self-attention and feed-forward network modules are order-agnostic to the input features, and the classic positional encoding \cite{vaswani2017attention} fails to capture graph structural information. To handle this problem, we develop dual-modality line graph transformers, which introduce efficient structural encoding methods to incorporate the topological and geometric structures, respectively. Figure \ref{transformer} illustrates the dual-modality line graph transformer architecture.

\subsubsection{2D Line Graph Transformer.} We introduce three distinct structural encoding methods: eigenvector positional encoding, path length encoding, and path node encoding, thus integrating the topological information of 2D line graphs into transformer architecture. These encoding methods serve as graph inductive bias to convey structural information, leading to more robust and discriminative graph representations.

\textit{Eigenvector Positional Encoding.} For the topological line graph, we use the eigenvectors of its normalized Laplacian matrix as positional encoding. Laplacian matrix is defined as the difference between the degree matrix and the adjacency matrix, representing the topology of a graph. Furthermore, Laplacian eigenvectors can distinguish the local position information of different nodes, while preserving the global topology structure \cite{dwivedi2020benchmarking, belkin2003laplacian}. For each node in the topological line graph, its corresponding eigenvector is viewed as a node positional encoding and added to the node feature for position-aware feature input:

\begin{equation}
    \mathbf{x}_{i}^{2d}= \mathbf{n}_i^{2d} + \mathbf{W}_v\mathbf{v}_i^{2d},
\end{equation}

\noindent where $\mathbf{v}_{i}^{2d} \in \mathbb{R}^K$ is the $K$ smallest non-trivial eigenvectors for the $i$-th node in the topological line graph, and $\mathbf{W}_v \in \mathbb{R}^{\Theta_{\hat{v}} \times K}$ denotes a learnable projection matrix.

\textit{Path Length Encoding.} Different from general structured data in Euclidean space, e.g., texts and images, graphs do not have a canonical order and can only lie in a non-Euclidean space connected by edges. To model full structural information, the spatial relations between nodes should be measured. Following the previous studies \cite{ying2021transformers, Li2022KPGTKP}, we derive the shortest path between each pair of nodes, and then leverage the path length and the node features along the path to capture spatial information. For the path length encoding, we assign a learnable embedding vector for each length scalar, then project it as a bias term in the self-attention module. Given a node pair $(\hat{v}_i, \hat{v}_j)$ in the topological line graph, its path length encoding can be written as:

\begin{equation}
  \mathbf{p}_{ij}=\mathrm{SPL}(\hat{v}_i, \hat{v}_j),
  b_{ij} = \mathbf{p}_{ij}^T\mathbf{w_p},
\end{equation}

\noindent where $\mathrm{SPL}(\cdot)$ denotes an embedding function that determines a learnable embedding vector $\mathbf{p}_{ij} \in \mathbb{R}^p$ indexed by the length of shortest path between node $\hat{v}_i$ and $\hat{v}_j$, and $\mathbf{w_p} \in \mathbb{R}^p$ is a learnable projection vector.

\textit{Path Node Encoding.} As the different nodes along the shortest path can distinguish spatial relations in graphs, we encode the path node features as additional structural information into modeling. For each node pair $(\hat{v}_i, \hat{v}_j)$,  we compute an average bias term of the dot-products of the node feature and a learnable projection vector along the path, which is defined as:

\begin{equation}
  \begin{split}
  c_{ij}&=\frac{1}{N_{ij}}\sum_{n=1}^{N_{ij}}(\mathbf{x}_n^{2d})^T\mathbf{w}_n,
  \end{split}
\end{equation}

\noindent where $N_{ij}$ is the shortest path length between node $\hat{v}_i$ and $\hat{v}_j$, $\mathbf{x}_n^{2d}$ is the $n$-th 2D node feature along the path, and $\mathbf{w}_n \in \mathbb{R}^{\Theta_{\hat{v}}}$ denotes the $n$-th learnable projection vector indexed by the path position.

Next, we incorporate the above three structural encodings into the transformer architecture, which captures full structural information in the topological line graph. Specifically, the eigenvector positional encoding is added to the original node feature as the input feature. The path length and path node encodings are viewed as two structural bias terms to the self-attention module. We can compute each element $(i, j)$ in the attention matrix as follows:

\begin{equation}
    \mathbf{A}_{ij}^{2d} = \frac{(\mathbf{W}_q^{2d}\mathbf{x}_i^{2d})^T(\mathbf{W}_k^{2d}\mathbf{x}_j^{2d})}{\sqrt{d_k}} + b_{ij} + c_{ij},
\end{equation}

\noindent where $\mathbf{W}^{2d}_q \in \mathbb{R}^{d_k \times \Theta_{\hat{v}}}$ and $\mathbf{W}^{2d}_k \in \mathbb{R}^{d_k \times \Theta_{\hat{v}}}$ are learnable projection matrices. Similar to the vanilla transformer, we augment the self-attention module by extending to the multi-head mechanism. Note that the structural bias terms $b_{ij}$ and $c_{ij}$ should also be computed by separate projection vectors within different heads. Then we adapt the feed-forward network with layer normalization and skip connection operations over the output of the self-attention module.

\subsubsection{3D Line Graph Transformer.} To incorporate the geometric graph information into transformer architecture, we consider two novel structural encoding methods: atomic distance encoding and geometric angle encoding. By leveraging the constructed geometric line graph, the atomic distance and geometric angle can be embedded into the node and edge representations in a natural manner. For the atomic distance encoding, we concatenate the atomic feature embedding and pair-wise distance embedding in Equation (\ref{atomic_dis}), which provides a relative position encoding as the input feature to the following modeling, that is, $\mathbf{x}_{i}^{3d} = \mathbf{n}_{uv}^{3d}$.

\textit{Geometric Angle Encoding.} As an invariant spatial feature, the geometric angle between the adjacent bonds plays a vital role in determining the spatial structure of molecules. Similarly, we derive the shortest path between any two nodes in the geometric line graph. Since the angle embedding is viewed as the line edge feature by Equation (\ref{angle}), we can encode the edge features along the path as the angle information. Given a node pair $(\hat{v}_i, \hat{v}_j)$, an average bias term is computed by taking the dot-product of the edge feature and a learnable projection vector along the shortest path:

\begin{equation}
  \begin{split}
  g_{ij}&=\frac{1}{N_{ij} - 1}\sum_{m=1}^{N_{ij} - 1}(\mathbf{e}_{m}^{3d})^T\mathbf{w}_{m},
  \end{split}
\end{equation}

\noindent where $N_{ij}$ is the shortest path length, $\mathbf{e}_{m}^{3d}$ is the $m$-th 3D edge feature along the path, and $\mathbf{w}_m \in \mathbb{R}^{\Theta_{\hat{e}}}$ denotes the $m$-th learnable projection vector indexed by the path position.

Then we introduce the pair-wise distance and angle encodings into transformer architecture, capturing the geometric structural information. For each element $(i, j)$ in the attention matrix of 3D line graph transformer, we have:

\begin{equation}
    \mathbf{A}_{ij}^{3d} = \frac{(\mathbf{W}_q^{3d}\mathbf{x}_i^{3d})^T(\mathbf{W}_k^{3d}\mathbf{x}_j^{3d})}{\sqrt{d_k}} + g_{ij},
\end{equation}

\noindent where $\mathbf{W}^{3d}_q \in \mathbb{R}^{d_k \times \Theta_{\hat{v}}}$ 
and $\mathbf{W}^{3d}_k \in \mathbb{R}^{d_k \times \Theta_{\hat{v}}}$ are learnable projection matrices. 
As with the 2D line graph transformer, we introduce the multi-head mechanism in the self-attention module, then feed the output into a feed-forward network with layer normalization and skip connection operations.

\vspace{-0.5em}
\subsection{Pre-training Task Construction}
We design two complementary tasks from both intra-modality and inter-modality levels: masked line node prediction and dual-view contrastive learning.

\noindent\textbf{Masked Line Node Prediction.} For the intra-modality level pre-training, we leverage the node masking to learn the regularities of the atom/bond attributes. Following the ``masked language model'' training objective in BERT \cite{devlin2018bert}, we randomly sample a proportion of line nodes in the topological and geometric line graphs, and replace them with the masking procedures: (i) masking 80\% of the selected nodes using a special indicator, (ii) replacing 10\% of the selected nodes with other random nodes, and (iii) keeping the remaining 10\% nodes unchanged. Then we apply 2D/3D line graph transformers to obtain the corresponding line node embeddings. Finally, a multi-layer perceptron (MLP) on top of the masked embeddings is used to predict the types of original nodes with a cross-entropy loss function. The mask losses on the 2D and 3D views are denoted as $\mathcal{L}^{2d}_{mask}$ and $\mathcal{L}^{3d}_{mask}$, respectively.

\noindent\textbf{Dual-view Contrastive Learning.} As the 2D and 3D views of the same molecule are correlated and provide complementary knowledge to each other, we can enhance molecular self-supervised learning by designing the inter-modality pre-training task. Specifically, we leverage contrastive learning to maximize the mutual information between the 2D and 3D representations of a molecule. In each line graph, we create a virtual node that is connected to all other nodes. Its output represents the graph-level embedding and can be used for inter-modality contrast. For each molecule, we first extract the graph-level embeddings from the masked topological and geometric line graphs, and then adapt the individual projection heads to map them into a common representation space, resulting in the derivation of $\mathbf{z}^{2d}_{i}$ and $\mathbf{z}^{3d}_{i}$. The 2D-3D latent vectors from the same molecule are regarded as positive pairs, and negative pairs otherwise. Based on the InfoNCE loss \cite{oord2018representation}, our dual-view contrastive learning is written as: 

\begin{equation}
    \mathcal{L}_{cl}^{2d} = -\sum_{i=1}^{N}\mathrm{log}\frac{\mathrm{exp}(\mathrm{sim}(\mathbf{z}^{2d}_{i}, \mathbf{z}^{3d}_{i})/\tau)}{\sum_{j=1}^{M}\mathrm{exp}(\mathrm{sim}(\mathbf{z}^{2d}_{i}, \mathbf{z}^{3d}_{j})/\tau)},
\end{equation}

\begin{equation}
    \mathcal{L}_{cl}^{3d} = -\sum_{i=1}^{N}\mathrm{log}\frac{\mathrm{exp}(\mathrm{sim}(\mathbf{z}^{3d}_{i}, \mathbf{z}^{2d}_{i})/\tau)}{\sum_{j=1}^{M}\mathrm{exp}(\mathrm{sim}(\mathbf{z}^{3d}_{i}, \mathbf{z}^{2d}_{j})/\tau)},
\end{equation}

\begin{equation}
    \mathcal{L}_{cl} = \frac{1}{N}(\mathcal{L}_{cl}^{2d} + \mathcal{L}_{cl}^{3d}),
\end{equation}

\noindent where $\mathbf{z}^{2d}_{i}$ and $\mathbf{z}^{3d}_{i}$ denotes two latent vectors in 2D and 3D views from the same molecule, $\tau$ is a temperature coefficient, $\mathrm{sim}(\cdot)$ measures the dot similarity between two vectors, $M$ is the batch size, and $N$ is the number of molecules in the dataset. Dual-view contrastive learning can simultaneously align the positive pairs of the same molecule and distinguish them from negative pairs, thereby enabling the 2D/3D representation to extract complementary information from its 3D/2D counterpart.

Finally, we combine the intra-modality mask losses and inter-modality contrastive loss into a complete objective function as follows:

\begin{equation}
    \mathcal{L}_{obj} = \lambda_1\mathcal{L}^{2d}_{mask} + \lambda_2\mathcal{L}^{3d}_{mask} + \lambda_3\mathcal{L}_{cl},
\end{equation}

\noindent where $\lambda_1$, $\lambda_2$ and $\lambda_3$ are weighting coefficients. The training objective is to minimize the overall loss $\mathcal{L}_{obj}$. 

\begin{table*}[t!]
  \centering
  \caption{Performance in AUROC on the downstream classification datasets (\textbf{Best}, \underline{Second Best}).}
\scalebox{0.8}{\begin{tabular}{l|cccccccccc}
\toprule
\multicolumn{1}{c|}{} & \multicolumn{9}{c}{Classification accuracy (AUROC) $\uparrow$} \\ 
\midrule
\multicolumn{1}{c|}{Dataset} & BBBP  & Sider  & ClinTox  & BACE & Tox21 & MUV & HIV & Estrogen  & MetStab  \\ 
\multicolumn{1}{c|}{\# Molecules} & 2,039  & 1,427  & 1,478  & 1,513 & 7,831 & 93,087 & 41,127 & 3,122  & 2,267  \\ 
\multicolumn{1}{c|}{\# Tasks} & 1  & 27  & 2  & 1 & 12 & 17 & 1 & 2  & 2  \\ 
\midrule
GraphCL  & $0.901_{(0.033)}$ & $0.625_{(0.023)}$ & $0.642_{(0.120)}$ & $0.868_{(0.025)}$ & $0.823_{(0.015)}$ & $0.762_{(0.027)}$ & $0.776_{(0.007)}$ & $0.891_{(0.020)}$ & $0.794_{(0.037)}$ \\
JOAO   & $0.900_{(0.031)}$ & $0.629_{(0.020)}$ & $0.707_{(0.068)}$  & $0.866_{(0.035)}$ & $0.825_{(0.011)}$ & $0.766_{(0.048)}$ & $0.769_{(0.004)}$ & $0.869_{(0.050)}$ & $0.814_{(0.031)}$ \\
GROVER  & $0.923_{(0.029)}$ & $0.645_{(0.010)}$ & $0.894_{(0.032)}$  & $0.882_{(0.030)}$  & $0.840_{(0.016)}$  & $\underline{0.831}_{(0.035)}$  & $0.773_{(0.014)}$  & $0.903_{(0.030)}$  & $0.822_{(0.039)}$  \\
3DInfomax  & $0.905_{(0.033)}$ & $0.634_{(0.029)}$ & $0.724_{(0.104)}$ & $0.862_{(0.023)}$ & $0.819_{(0.021)}$  & $0.803_{(0.007)}$ & $0.750_{(0.010)}$ & $0.871_{(0.039)}$ & $0.811_{(0.043)}$ \\
GraphMVP & $0.918_{(0.019)}$ & $0.652_{(0.027)}$  & $0.705_{(0.092)}$  & $0.866_{(0.028)}$ & $0.832_{(0.017)}$ &  $0.787_{(0.039)}$ &$\underline{0.791}_{(0.004)}$  & $0.892_{(0.033)}$  & $0.827_{(0.043)}$  &  \\
KPGT  & $\underline{0.927}_{(0.028)}$  & $\underline{0.658}_{(0.013)}$ & $\textbf{0.915}_{(0.027)}$ & $\underline{0.893}_{(0.032)}$ & $\underline{0.847}_{(0.013)}$ & $0.829_{(0.047)}$ & $0.768_{(0.007)}$ & $\underline{0.915}_{(0.028)}$ & $\underline{0.846}_{(0.052)}$
\\
\midrule
Galformer & $\textbf{0.933}_{(0.027)}$ & $\textbf{0.681}_{(0.011)}$ & $\underline{0.910}_{(0.030)}$ & $\textbf{0.897}_{(0.026)}$ & $\textbf{0.852}_{(0.015)}$ & $\textbf{0.841}_{(0.023)}$ &$\textbf{0.796}_{(0.017)}$ & $\textbf{0.936}_{(0.010)}$ & $\textbf{0.875}_{(0.029)}$ \\ 
\bottomrule
\end{tabular}}
\label{downstream_result_class}

\vspace{-1em}

\end{table*}

\section{Experiments}

\subsection{Experimental Settings}
\textbf{Benchmark Datasets.} For pre-training, we use the GEOM \cite{axelrod2022geom} dataset comprising 304k molecules, which contains precise 2D and 3D structures. Most molecules have multiple 3D conformers, but the conformer with the lowest energy is deemed the most stable and has the highest possibility. Therefore, we take the lowest-energy conformer of each molecule as its 3D input. For downstream fine-tuning, we evaluate model performance on twelve labeled 2D molecular datasets from MoleculeNet \cite{wu2018moleculenet} and other public sources \cite{gaulton2012chembl, podlewska2018metstabon, gamo2010thousands, hachmann2011harvard}, including nine classification tasks and three regression tasks. These benchmarks provide comprehensive coverage of molecular properties, including physiology, biophysics and physical chemistry.

\noindent\textbf{Data Splitting.} As suggested in MoleculeNet \cite{wu2018moleculenet}, we adopt scaffold split to create a challenging yet realistic evaluation for molecular property prediction. Each downstream dataset is divided into training/validation/test with an 8:1:1 ratio. We conduct three independent scaffold splitting runs with different random seeds, and report the means and standard deviations of evaluation metrics.

\noindent\textbf{Baselines.} We comprehensively evaluate the performance of Galformer against six state-of-the-art self-supervised learning methods on molecules. Among these, GraphCL \cite{You2020GraphCL} and JOAO \cite{you2021graph} are 2D contrastive methods that leverage data augmentation strategies on 2D molecular graphs to facilitate representation learning. GROVER \cite{rong2020self} and KPGT \cite{Li2022KPGTKP} are 2D generative methods that incorporate domain knowledge for the design of self-supervised learning tasks. 3DInformax \cite{Strk20213DII} and GraphMVP \cite{liu2022pretraining} are 2D-3D pre-training methods with GNN-based backbones. 3DInformax aims to maximize the mutual information between the 2D and 3D representations via contrastive learning, while GraphMVP adopts a hybrid strategy to learn complementary 2D-3D information.

\noindent\textbf{Implementation Details.} We mainly implement Galformer in PyTorch. The Adam optimizer with a polynomial decay learning rate scheduler is used for training optimization. We employ 12-layer 2D and 3D line graph transformers as backbone networks. The hidden size is set to 768 and the number of attention heads is set to 12. We select a batch size of 256 and pre-train the model for 50 epochs. The peak learning rate is set to 2e-4 and Adam weight decay is set to 1e-6. Additionally, we set the mask ratio as 0.4 and temperature coefficient $\tau$ as 0.1. The multi-task weighting coefficients $\lambda_1$, $\lambda_2$ and $\lambda_3$ are all set to 1 for balanced optimization. After pre-training, we concatenate the graph-level embedding and averaged line node embedding as the molecular representation for downstream tasks.

In the process of downstream fine-tuning, we use a two-layer MLP with ReLU activation as the prediction head, which is added on top of the pre-trained 2D backbone network. The binary cross-entropy and root mean square error (RMSE) losses are adapted for classification and regression tasks, respectively. The hidden dimension of prediction head is set to 256 and the batch size is set to 32. A grid search for hyper-parameters is performed, exploring 6 combinations of learning rates [3e-5, 1e-5, 3e-6] and weight decays [0, 1e-6] to identify the optimal choice for each task. We fine-tune 50 epochs and report the testing score with the model of best performing epoch on the validation set. All experiments are conducted on Tesla A100 GPUs.

\noindent\textbf{Evaluation Metrics.} We use two metrics for the performance evaluation of downstream datasets: the area under the receiver operating characteristic curve (AUROC) for classification datasets, and RMSE for regression datasets.

\subsection{Performance Evaluation}

\begin{table}[t]
  \centering
  \caption{Performance in RMSE on the downstream regression datasets (\textbf{Best}, \underline{Second Best}).}
\scalebox{0.85}{\begin{tabular}{l|ccc}
\toprule
\multicolumn{1}{c|}{} & \multicolumn{3}{c}{Regression error (RMSE) $\downarrow$} \\
\midrule
\multicolumn{1}{c|}{Dataset} & ESOL  & Malaria  & CEP \\
\multicolumn{1}{c|}{\# Molecules} & 1,128  & 9,999  & 29,978 \\
\multicolumn{1}{c|}{\# Tasks} & 1  & 1  & 1 \\ 
\midrule
GraphCL & $1.253_{(0.023)}$  & $1.115_{(0.127)}$ & $1.251_{(0.029)}$ \\
JOAO  & $1.203_{(0.061)}$ & $1.117_{(0.122)}$ & $1.278_{(0.021)}$ \\
GROVER & $0.928_{(0.106)}$ & $1.092_{(0.086)}$  & $1.064_{(0.085)}$ \\
3DInfomax  & $1.167_{(0.148)}$ & $\underline{1.075}_{(0.103)}$ & $1.277_{(0.019)}$ \\
GraphMVP  & $0.995_{(0.113)}$ & $1.101_{(0.107)}$ & $1.248_{(0.010)}$ \\
KPGT  & $\underline{0.802}_{(0.096)}$ & $1.083_{(0.082)}$ & $\underline{1.020}_{(0.100)}$ \\
\midrule
Galformer & $\mathbf{0.741}_{(0.087)}$ & $\mathbf{1.066}_{(0.092)}$ & $\mathbf{1.015}_{(0.084)}$ \\ 
\bottomrule
\end{tabular}
}
\vspace{-1.5em}
\label{downstream_result_reg}
\end{table}

\textbf{Overall Comparison.} Tables \ref{downstream_result_class} and \ref{downstream_result_reg} present the testing performance of all methods on classification and regression datasets, respectively. The results provide the following observations: (1) Galformer consistently achieves the best performance on 11 out of 12 downstream datasets. Compared to previous SOTA results from baselines, Galformer has an overall relative improvement of 1.3\% on the classification tasks and 2.6\% on the regression tasks. (2) Among contrastive learning baselines, GraphMVP and 3DInformax, which leverage both 2D and 3D graphs in pre-training, perform better on the whole average than GraphCL and JOAO that only use 2D graph. It indicates that incorporating 3D geometry can facilitate molecular self-supervised learning. Furthermore, Galformer achieves more significant improvement compared with the 2D-3D contrastive baselines. This result can be attributed to our complementary pre-training tasks, which can adaptively capture both inter-modality and intra-modality information. (3) GROVER and KPGT outperform other baselines for most tasks. One potential explanation is that they both employ transformer-based architectures as backbone networks, leading to higher model capacity and more expressive power than message passing models. Nevertheless, the lack of leveraging 3D geometry limits their ability to fully learn molecular representations. Galformer exhibits a notable capability to surpass the inherent limitations in previous methods, consequently achieving more accurate molecular property prediction.

\noindent\textbf{Ablation Studies.} We conduct ablation studies to analyze the effectiveness of the key components in Galformer.

\begin{figure}[t]
    \centering
    \includegraphics[width=0.42\textwidth]{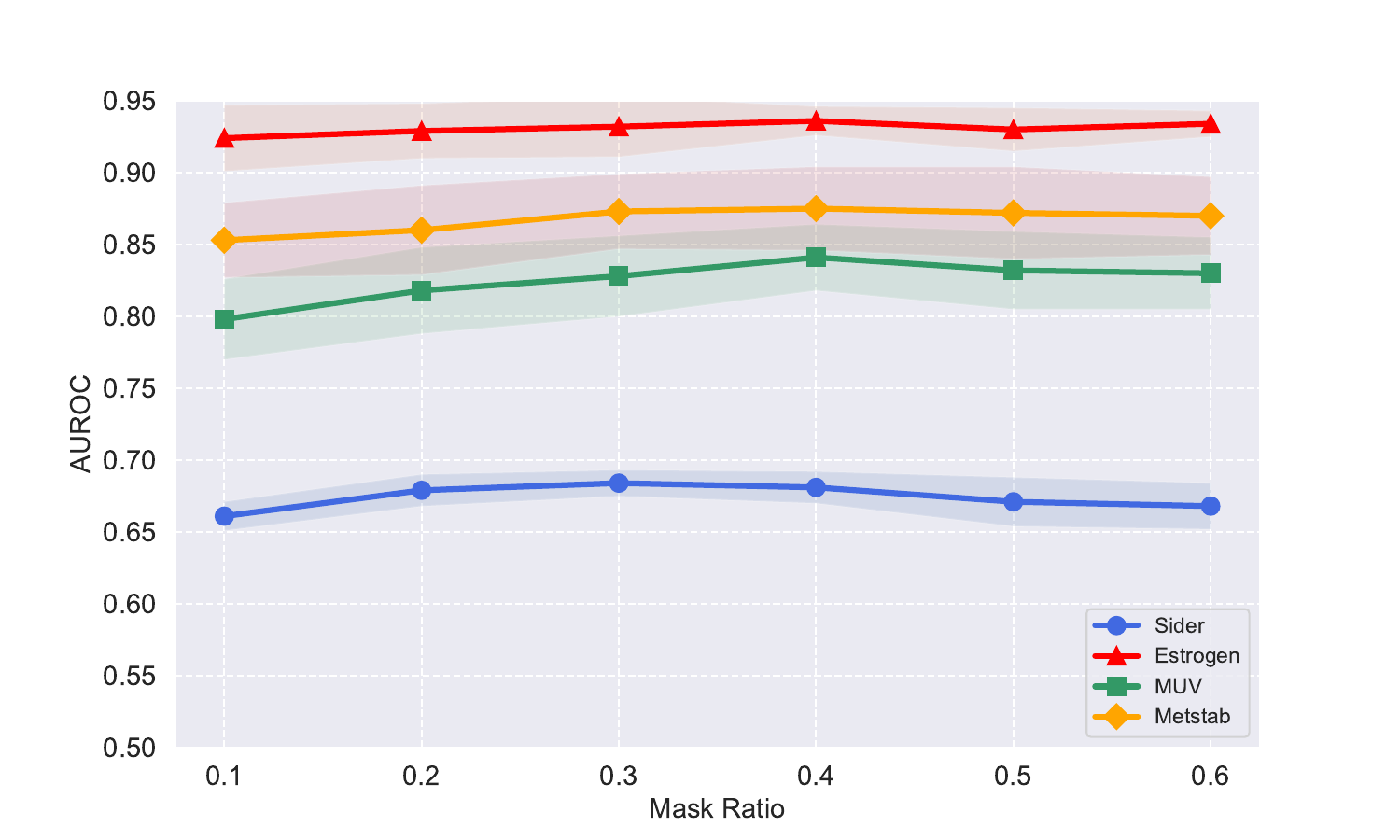}
    \caption{Performance variation of Galformer over different mask ratios on four benchmark datasets.} 
    \label{mask_ratio}
    \vspace{-1.5em}
\end{figure}

\textit{Effect of Pre-training Strategy and Backbone Network.} To verify the effectiveness of our pre-training strategy, we compare Galformer to the model without pre-training. Table \ref{table:backbones} shows that pre-trained Galformer yields an average improvement of 5.9\% in AUROC and 10.2\% in RMSE, demonstrating the effectiveness of our pre-training strategy. Next, we create two variants by replacing the 2D backbone network of Galformer with vanilla transformer \cite{vaswani2017attention} and Graphormer \cite{ying2021transformers}, respectively. The results in Table \ref{table:backbones} indicate that Galformer still performs the best, showing the efficacy of our line graph transformer backbone. This improvement can be attributed to our graph structural encodings that capture crucial structures in molecules.

\begin{table}[htbp]
  \centering
  \caption{Ablation study with the absence of pre-training and different backbones (\textbf{Best}, \underline{Second Best}).}
  \scalebox{0.85}{\begin{tabular}{l|cc}
  \toprule
  \makecell{Backbone}  & \makecell{Classification \\ \textbf{Avg. AUROC} $\uparrow$} & \makecell{Regression \\ \textbf{Avg. RMSE} $\downarrow$} \\ \midrule
  No Pre-training        & 0.810         & 1.048  \\ 
  Vanilla Transformer
  & 0.825         & 1.012  \\
  Graphormer 
  & \underline{0.854}         & \underline{0.951}  \\
  Galformer          & \textbf{0.858}         & \textbf{0.941}  \\ \bottomrule
  \end{tabular}
  }
  \label{table:backbones}
  \vspace{-2.0em}
\end{table}

\begin{table}[htbp]
  \centering
  \caption{Ablation study of the different contrastive losses (\textbf{Best}, \underline{Second Best}).}
\scalebox{0.85}{\begin{tabular}{l|cc}
  \toprule
  \multicolumn{1}{c|}{\makecell{Contrastive Loss (CL)}}  & \makecell{Classification \\ \textbf{Avg. AUROC} $\uparrow$} & \makecell{Regression \\ \textbf{Avg. RMSE} $\downarrow$} \\ \midrule
  w/o CL              & 0.838         & 0.986           \\
  NT-Xent
  & 0.850         & \underline{0.954}                   \\
  EMB-NCE
  & \underline{0.852}         & 0.956                   \\
  InfoNCE
  & \textbf{0.858}         & \textbf{0.941}            \\ 
  \bottomrule
  \end{tabular}
  }
  \label{table:cl_loss}
  \vspace{-0.5em}
\end{table}

\textit{Mask Ratio Analysis.} We investigate the impact of different mask ratios on the model performance, as depicted in Figure \ref{mask_ratio}. Galformer has the best overall performance with a mask ratio of 0.4. It is higher than the commonly employed ratio in previous works \cite{devlin2018bert, hu2020pretraining, Wang2021MolecularCL}, which were typically below 0.25. As our additional contrastive learning provides complementary knowledge from different modalities, Galformer can adapt to more challenging tasks with a higher mask ratio. This shares similarity with the success of masked autoencoders \cite{he2022masked} in the vision domain, where a high mask ratio can better capture long-range dependencies and improve model generalization.

\textit{Choice of Contrastive Loss.} To validate the effectiveness of the contrastive loss, we implement three variants: one without contrastive learning, and two others with NT-Xent \cite{chen2020simple} and EMB-NCE \cite{liu2022pretraining} contrastive losses, respectively. As observed from Table \ref{table:cl_loss}, applying contrastive loss improves performance on both classification and regression tasks. Furthermore, InfoNCE consistently outperforms the other variants, which verifies its effectiveness in cross-modality contrastive learning.

\begin{figure}[t]
    \centering
    \includegraphics[width=0.48\textwidth]{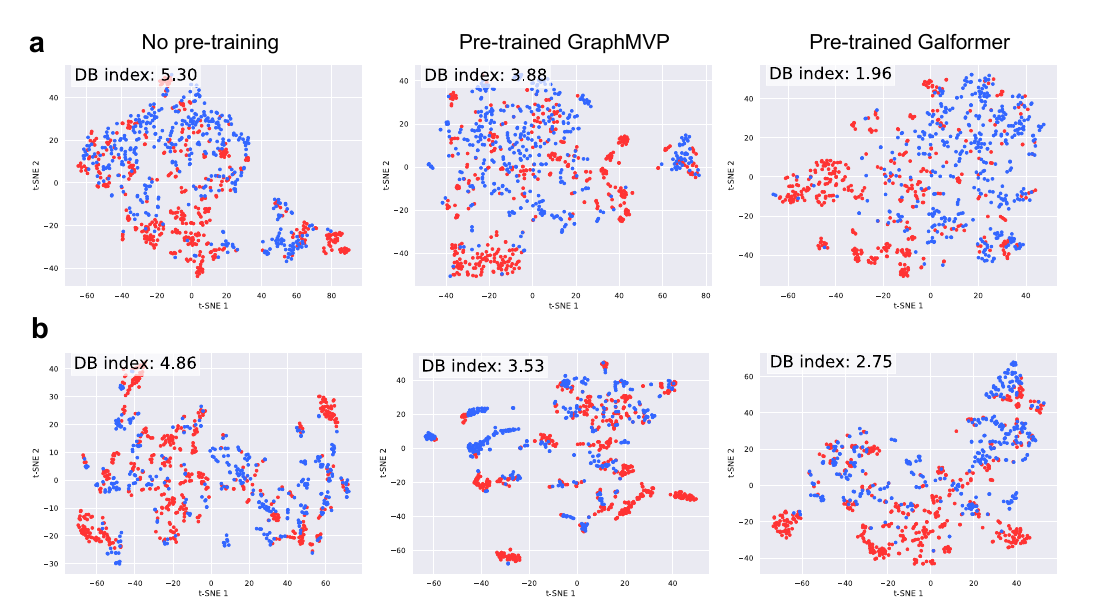}
    \caption{Visualization of the pre-trained molecular representations via $t$-SNE. (a) BBBP dataset. (b) ESOL dataset. The lower the DB index, the better the separation.}
    \label{visual}
    \vspace{-1.5em}
\end{figure}

\noindent\textbf{Molecular Representation Visualization.} To intuitively investigate the effect of pre-trained molecular representations without fine-tuning, we randomly select 800 molecules from BBBP and ESOL datasets respectively, and utilize $t$-SNE to project these representations onto a 2D embedding space for visualization. 
On BBBP, each molecule is associated with a binary class label, indicating its barrier permeability. 
On ESOL, the original label is a continuous variable that denotes the water solubility of the molecule.
We use the median 0.057 mols/L in ESOL as a binary threshold. Figure \ref{visual} shows the visualization results, using the Davies Bouldin (DB) index \cite{davies1979cluster} to measure the cluster separation. 
The lower the DB index, the better the separation. We can observe that pre-trained GraphMVP and Galformer can better separate the labeled molecules than the model without pre-training. Furthermore, Galformer decreases the DB index to 1.96 on BBBP, compared with 3.88 of GraphMVP. The results indicate that Galformer can effectively incorporate generalizable domain knowledge during pre-training, leading to advantages in downstream tasks.
\section{Conclusion}
In this work, we propose Galformer, a dual-modality pre-training framework for molecular property prediction. Extensive experiments on downstream tasks demonstrate its effectiveness over SOTA baselines. There are two potential future directions to extend: (1) Simulating the process of molecule generation and adapting Galformer in structure optimization. (2) Exploring the combination of Galformer with reinforcement learning to further improve its generalization.

\bibliography{aaai24}

\end{document}